\documentclass[12pt,a4paper,notitlepage,oneside]{article}

\usepackage{booktabs}
\usepackage{tcolorbox}
\usepackage{longtable}
\usepackage{hyperref}
\usepackage[capposition=top]{floatrow}
\hypersetup{colorlinks,allcolors=red}

\usepackage{listings}
\usepackage{xcolor}
\definecolor{codegreen}{rgb}{0,0.6,0}
\definecolor{codegray}{rgb}{0.5,0.5,0.5}
\definecolor{codepurple}{rgb}{0.58,0,0.82}
\definecolor{backcolour}{rgb}{0.95,0.95,0.92}

\lstdefinestyle{mystyle}{backgroundcolor=\color{backcolour}, commentstyle=\color{codegreen},
    keywordstyle=\color{magenta}, numberstyle=\tiny\color{codegray}, stringstyle=\color{codepurple}, basicstyle=\ttfamily\footnotesize,  breakatwhitespace=false, breaklines=true, captionpos=t, keepspaces=true, numbers=left,  numbersep=5pt,showspaces=false, showstringspaces=false, showtabs=false, tabsize=2}
\usepackage[backend=biber,style=numeric,sorting=none]{biblatex}
\begin{document}

\newcommand{\R}{\texttt{R}}
\newcommand{\Rs}{\texttt{R }}
\newcommand{\py}{\texttt{python}}
\newcommand{\pys}{\texttt{python }}
\newcommand{\SSs}{\texttt{SurvSet }}
\newcommand{\SSn}{\texttt{SurvSet}}

\title{\texttt{SurvSet}: An open-source time-to-event dataset repository}

\author{Erik Drysdale\thanks{erikinwest@gmail.com}}

\maketitle

\begin{abstract}
	Time-to-event (T2E) analysis is a branch of statistics that models the duration of time it takes for an event to occur. Such events can include outcomes like death, unemployment, or product failure. Most modern machine learning (ML) algorithms, like decision trees and kernel methods, are supported for T2E modelling with data science software (\pys and \R). To complement these developments, \SSs is the first open-source T2E dataset repository designed for a rapid benchmarking of ML algorithms and statistical methods. The data in \SSs have been consistently formatted so that a single preprocessing method will work for all datasets. \SSs currently has 76 datasets which vary in dimensionality, time dependency, and background (the majority of which come from biomedicine). \SSs is available on \href{https://pypi.org/project/SurvSet/}{\texttt{PyPI}} and can be installed with \texttt{pip install SurvSet}. \texttt{R} users can download the data directly from the corresponding \href{https://github.com/ErikinBC/SurvSet/tree/main/SurvSet/_datagen/output}{git repository}.
\end{abstract}

\section{Introduction \label{sec:intro}}
Many disciplines study phenomena in which a certain amount of time must pass before an event occurs. In statistics, this is referred to as time-to-event (T2E) analysis. Such processes are usually characterized by two properties: i) non-negative values, and ii) right-censoring. The latter refers to an observation where the event has not yet occurred. Implicit in T2E processes is that with sufficient measurement time, an event will inevitably occur. T2E analysis frequently arises in fields like biomedicine (death or relapse), economics (unemployment), e-commerce (customer churn), and engineering (product failure). T2E methods are designed to account for both right-censoring and non-negative values.

Due to the preponderance of biomedical research which studies end-points like death, T2E methods are often referred to as survival analysis. Both terms will be used interchangeably throughout this paper. Classical survival analysis is a well-developed field with mature computational resources. More recently, core machine learning (ML) algorithms have been adapted for T2E modelling, with some examples shown below. For a comprehensive survey of survival methods in ML see \cite{wang2019}.

\begin{itemize}
    \setlength\itemsep{-1mm}
    \item Regularized linear models (e.g. elastic net \cite{simon2011})
    \item Decision trees \cite{ciampi1987}
    \item Ensemble methods (e.g. random forests \cite{ishwaran2008} and gradient boosters \cite{hothorn2006})
    \item Kernel methods (e.g. SVMs \cite{van2011})
    \item Deep learning (e.g. feed-forward neural networks \cite{kvamme2019})
\end{itemize}

For most researchers and practitioners, the actual implementation of survival methods (both classical and ML-based) is done with special packages in \Rs and \py. The former includes packages like \texttt{survival} \cite{therneau2015} and \texttt{mlr3proba} \cite{sonabend2021} while the latter includes \texttt{lifelines} \cite{davidson2019}, \texttt{pysurvival} \cite{pysurvival}, and \texttt{scikit-survival} \cite{polsterl2020}. These packages will often have a handful of datasets to allow for testing and benchmarking.

\SSs seeks to complement these modelling packages by providing the first-ever open-source T2E dataset repository for the benchmarking and evaluation of ML algorithms and statistical methods. Benchmarking and contests have played an important role in the development and refinement of ML algorithms  \cite{caruana2006}. \SSs is comprised of 76 datasets that are structured in a consistent format. This enables a single preprocessing method that works for all datasets, and removes the need for researchers to spend time curating and formatting datasets. The datasets also vary in dimensionality, background, and whether the covariates vary across time.\footnote{The origins of this dataset were for testing regularity conditions of the False Positive Control Lasso \cite{drysdale2019}.}

\section{Overview \& design \label{sec:overview}}
\subsection{Usage}

\SSs is built around a single class, \texttt{SurvLoader}, which loads datasets through the \texttt{load\_dataset} method. This method is a convenient wrapper for loading the underlying comma-separated files. As was previously mentioned, these files can be accessed directly through \href{https://github.com/ErikinBC/SurvSet/tree/main/SurvSet/\_datagen/output}{github}. The code block in \ref{lst:ex} provides an example of how to load a dataset with \SSn. The full list of available datasets can be seen in the attribute \texttt{df\_ds} (see Table \ref{tab:df_ds}). 

The \texttt{load\_dataset} method returns a dictionary with keys \texttt{df} for the underlying DataFrame (see Table \ref{tab:df_ex}) and a URL with a further description on the columns.\footnote{See, for example, the \href{https://rdrr.io/cran/CoxRidge/man/ova.html}{\texttt{ova}} reference.} The columns of each dataset are consistently structured. The observation ID, event, and survival times are placed in the first three (or four) columns. The remaining feature columns have a prefix to indicate whether they are categorical and can be one-hot-encoded, or are numeric and can be normalized. The column order and names are indicated below.

\begin{enumerate}
    \setlength\itemsep{-1mm}
    \item \texttt{pid}: the unique observation identifier (relevant for datasets with time-varying features)
    \item \texttt{event}: a binary event indicator (1=event has happened) 
    \item \texttt{time}: time to event/censoring (or start time if \texttt{time2} exists)
    \item \texttt{time2}: end time $[\texttt{time}, \texttt{time2})$ if there are time-varying features (non-existent otherwise)
    \item \texttt{num\_\{\}}: prefix implies a continuous feature
    \item \texttt{fac\_\{\}}: prefix implies a categorical feature
\end{enumerate}

\bigbreak

\begin{lstlisting}[language=Python, caption={Example of loading data \label{lst:ex}}]
from SurvSet.data import SurvLoader
loader = SurvLoader()
# List of available datasets and meta-info
print(loader.df_ds.head())
# Load dataset and its reference
df, ref = loader.load_dataset(ds_name=`ova`).values()
print(df.head())
\end{lstlisting}

\begin{longtable}[ht]{llllll}
    \caption{Example of dataset list: \texttt{df\_ds} \label{tab:df_ds}}
    \\ \hline 
    ds\_name & is\_td & n & n\_fac & n\_ohe & n\_num \\
    \hline
    hdfail & False & 52422 & 5 & 87 & 1 \\
    stagec & False & 146 & 4 & 15 & 3 \\
    veteran & False & 137 & 3 & 5 & 3 \\
    vdv & False & 78 & 0 & 0 & 4705 \\
    AML\_Bull & False & 116 & 0 & 0 & 6283 \\
    \hline
\end{longtable}

\begin{longtable}[ht]{llllllll}
    \caption{Example of \texttt{ova} dataset: \texttt{df} \label{tab:df_ex}}
    \\ \hline 
    pid & event & time & num\_karn & num\_diam & fac\_karn & fac\_diam &
    fac\_figo \\
    \hline
    1 & 1 & 7 & 1 & 3 & 1 & 3 & 1 \\
    2 & 1 & 8 & 2 & 2 & 2 & 2 & 1 \\
    3 & 1 & 9 & 4 & 4 & 4 & 4 & 1 \\
    4 & 1 & 10 & 4 & 4 & 4 & 4 & 1 \\
    5 & 1 & 13 & 2 & 4 & 2 & 4 & 1 \\
    \hline
\end{longtable}

\subsection{Datasets}

\begin{figure}[htbp]
    \centering
    \caption{\SSs datasets and dimensionality \label{fig:gg_ds}}
    \includegraphics[scale=0.85]{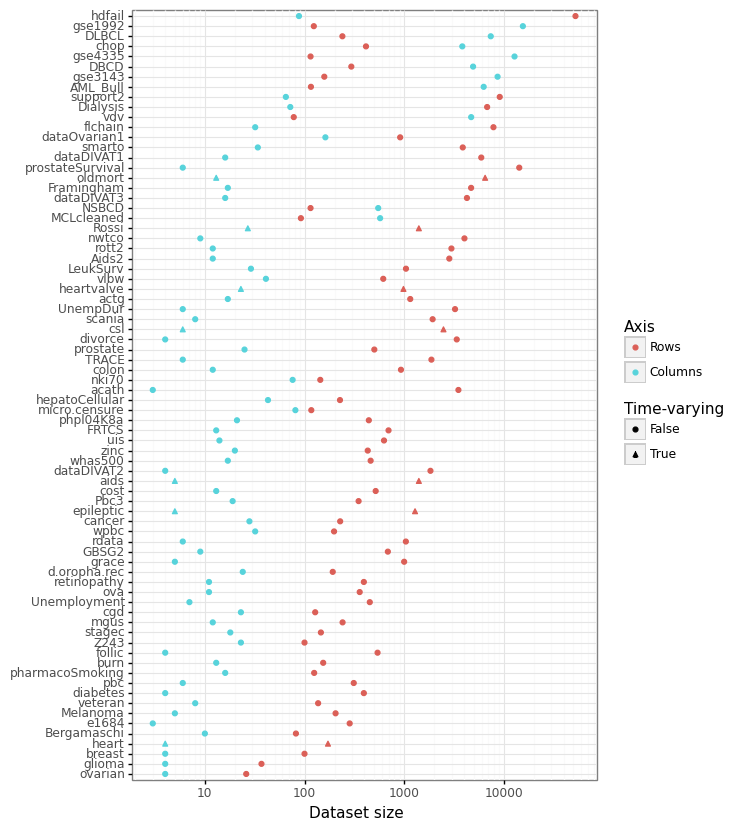}
\end{figure}

\begin{figure}[htbp]
    \centering
    \caption{Concordance of a regularized linear model on \SSs \label{fig:gg_cindex}}
    \includegraphics[scale=0.85]{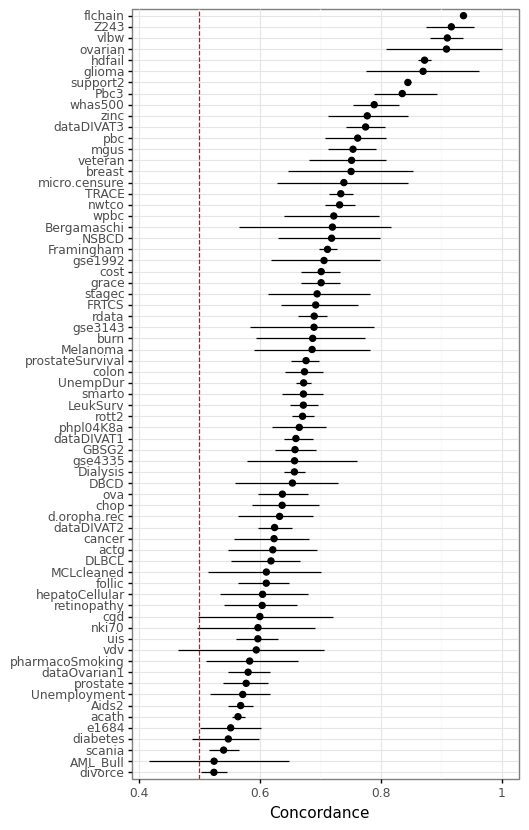}
    \floatfoot{Based on random 30\% test set; uses \texttt{CoxnetSurvivalAnalysis} from \texttt{scikit-survival} with default settings; concordance is equivalent to c-index \cite{harrell1982}}
\end{figure}

\SSs currently has 76 datasets which vary in dimensionality (see Figure \ref{fig:gg_ds}). This includes high-dimensional genomics datasets ($p \gg n$) like \texttt{gse1992}, and long and skinny datasets like \texttt{hdfail} ($n \gg p$). Most of these datasets come from existing \Rs packages, although not exclusively (see Table \ref{tab:ref_table}). An initial empirical experiment suggests that fitting a (regularized) linear model to each dataset produces a roughly uniform distribution of concordance results (see Figure \ref{fig:gg_cindex}).

The construction of these datasets was necessarily subjective as decisions needed to be made about which columns were relevant and which columns needed to dropped (especially if they had information leakage). For the datasets with time-varying features, time intervals were aggregated so that the minimum number of rows were needed to capture all information about changes in features. For example, if a measurement was made every 10 minutes, but the first change of a feature occured at the 50 minute mark, then the first interval would span [0,50) minutes. Users who are interested in how each dataset was curated can explore the processing files in the \texttt{\_datagen} folder (which otherwise should not be used).

\section{Conclusion \label{sec:conclusion}}
\SSs is a powerful resource to enable improvements for ML algorithms in T2E analysis. Currently, novel algorithms seen in the literature are only ever benchmarked against a handful of datasets. Furthermore, these datasets will vary by project making it difficult to assess what constitutes state-of-the-art performance. The hope is that \SSs will become a well-known resource for the ML community that focuses on survival analysis. This package can easily be expanded with new datasets and interested parties are encouraged to provide suggestions and contribute.

\break
\begin{center}
    \begin{longtable}{c|c|c|c}
        \caption{Dataset origin \label{tab:ref_table}} \\
         Dataset & Package & Data source & Package  \\
         \hline\hline
         \texttt{MCLcleaned} & \texttt{AdapEnetClass} & \cite{rosenwald2003} & \cite{AdapEnetClass} \\ 
         \hline
         \texttt{rott2} & \texttt{AF} & \cite{royston2011}  & \cite{AF} \\ 
         \hline
         \texttt{hepatoCellular} & \texttt{asaur} & \cite{li2014}  &  \cite{asaur} \\
         \texttt{pharmacoSmoking} & \texttt{asaur} & \cite{steinberg2009}  &  \\ 
         \texttt{prostateSurvival} & \texttt{asaur} & \cite{lu2009}  &  \\ 
         \hline
         \texttt{chop} & \texttt{bujar} & \cite{lenz2008}  & \cite{bujar} \\ 
         \hline
         \texttt{glioma} & \texttt{coin} & \cite{grana2002}  & \cite{coin} \\ 
         \hline
         \texttt{ova} & \texttt{CoxRidge} & \cite{van1989,verweij1993}  & \cite{CoxRidge} \\ 
         \hline
         \texttt{breast} & \texttt{coxphf} & \cite{losch1998,heinze2001}  & \cite{coxphf} \\ 
         \hline
         \texttt{UnempDur} & \texttt{ecdat} & \cite{mccall1996}  & \cite{ecdat} \\
         \texttt{Unemployment} & \texttt{ecdat} & \cite{romeo1999}  &  \\ 
         \hline
         \texttt{scania} & \texttt{eha} & \cite{dribe2020}  & \cite{eha} \\
         \texttt{oldmort} & \texttt{eha} & \cite{edvinsson2000}  &  \\ 
         \hline
         \texttt{hdfail} & \texttt{frailtySurv} & \cite{backblaze}  & \cite{frailtySurv} \\ 
         \hline
         \texttt{uis} & \texttt{Hosmer} & \cite{hosmer2002}  & \cite{hosmer2002} \\ 
        \texttt{FRTCS} & \texttt{Hosmer} & \cite{hosmer2002}  &  \\          
         \hline
         \texttt{smarto} & \texttt{hdnom} & \cite{simons1999}  & \cite{hdnom} \\ 
         \hline
         \texttt{burn} & \texttt{iBST} & \cite{ichida1993}  & \cite{iBST} \\ 
         \hline
         \texttt{d.oropha.rec} & \texttt{invGauss} & \cite{kalbfleisch2011}  & \cite{invGauss} \\ 
         \hline
         \texttt{aids} & \texttt{JM} & \cite{goldman1996}  & \cite{JM} \\
         \hline
         \texttt{heartvalve} & \texttt{joineR} & \cite{lim2008}  & \cite{joineR} \\
         \texttt{epileptic} & \texttt{joineR} & \cite{marson2007}  &  \\ 
         \hline
         \texttt{dataOvarian1} & \texttt{joint.Cox} & \cite{ganzfried2013}  & \cite{jointCox} \\ 
         \hline
         \texttt{aids2} & \texttt{MASS} & \cite{ripley1994}  & \cite{MASS} \\ 
         \texttt{melanoma} & \texttt{MASS} & \cite{drzewiecki1985,drzewiecki1980}  & \\
         \hline
         \texttt{grace} & \texttt{mlr3proba} & \cite{tang2007, hosmer2002}  & \cite{mlr3proba} \\
         \texttt{actg} & \texttt{mlr3proba} & \cite{hammer1997, hosmer2002}  & \\ 
         \hline
         \texttt{phpl04K8a} & \texttt{openml} & \cite{shedden2008}  & \cite{OpenML} \\ 
         \hline
         \texttt{zinc} & \texttt{NestedCohort} & \cite{abnet2004}  & \cite{NestedCohort} \\ 
         \hline
         \texttt{Pbc3} & \texttt{pec} & \cite{lombard1993}  & \cite{pec} \\
         \texttt{cost} & \texttt{pec} & \cite{jorgensen1996}  & \\ 
         \texttt{GBSG2} & \texttt{pec} & \cite{schumacher1994}  & \\ 
         \hline
         \texttt{nki70} & \texttt{penalized} & \cite{van2002}  & \cite{penalized} \\ 
         \hline
         \texttt{micro.censure} & \texttt{plsRcox} & \cite{romain2010}  & \cite{plsRcox} \\ 
         \hline
         \texttt{divorce} & \texttt{princeton} & \cite{lillard2003}  & \cite{princeton} \\ 
         \hline
         \texttt{follic} & \texttt{randomForestSRC} & \cite{petersen2004}  & \cite{randomForestSRC} \\
         \texttt{vdv} & \texttt{randomForestSRC} & \cite{vant2002}  & \\ 
         \hline
         \texttt{Bergamaschi} & \texttt{RCASPAR} & \cite{bergamaschi2006}  & \cite{RCASPAR} \\ 
         \hline
         \texttt{Dialysis} & \texttt{RcmdrPlugin.survival} & \cite{sa2003}  & \cite{RcmdrPluginsurvival} \\
         \texttt{Rossi} & \texttt{RcmdrPlugin.survival} & \cite{rossi1980}  & \\ 
         \hline
         \texttt{rdata} & \texttt{relsurv} & \cite{pohar2006}  & \cite{relsurv} \\ 
         \hline
         \texttt{NSBCD} & \texttt{Reddy} & \cite{sorlie2003}  & \cite{Reddy}  \\ 
         \texttt{AML Bull} & \texttt{Reddy} & \cite{bullinger2004} & \\ 
         \texttt{DBCD} & \texttt{Reddy}  & \cite{van2006} & \\ 
         \texttt{DLBCL} & \texttt{Reddy} & \cite{rosenwald2002} & \\ 
         \hline
         \texttt{DIVAT1} & \texttt{RISCA} & \cite{divat2022}  & \cite{RISCA} \\ 
         \texttt{DIVAT2} & \texttt{RISCA} & \cite{divat2022,le2016}  & \\ 
         \texttt{DIVAT3} & \texttt{RISCA} & \cite{divat2022,le2018}  & \\ 
         \hline
         \texttt{Z243} & \texttt{RobustAFT} & \cite{marazzi1993}  & \cite{RobustAFT} \\ 
         \hline
         \texttt{stagec} & \texttt{rpart} & \cite{breiman2017}  & \cite{rpart} \\ 
         \hline
         \texttt{e1684} & \texttt{smcure} & \cite{kirkwood1996}  & \cite{smcure} \\ 
         \hline
         \texttt{whas500} & \texttt{smoothHR} & \cite{hosmer2002}  & \cite{smoothHR} \\ 
         \hline
         \texttt{LeukSurv} & \texttt{spBayesSurv} & \cite{henderson2002}  & \cite{spBayesSurv} \\ 
         \hline
         \texttt{cancer} & \texttt{survival} & \cite{loprinzi1994}  & \cite{survival} \\ 
         \texttt{cgd} & \texttt{survival} & \cite{international1991,fleming2011}  & \\
         \texttt{colon} & \texttt{survival} & \cite{moertel1995}  & \\
         \texttt{flchain} & \texttt{survival} & \cite{kyle2006,dispenzieri2012}  & \\
         \texttt{heart} & \texttt{survival} & \cite{crowley1977}  & \\
         \texttt{mgus} & \texttt{survival} & \cite{kyle1993}  & \\
         \texttt{ovarian} & \texttt{survival} & \cite{edmonson1979}  & \\
         \texttt{pbc} & \texttt{survival} & \cite{dickson1989}  & \\
         \texttt{retinopathy} & \texttt{survival} & \cite{blair1980}  & \\
         \texttt{veteran} & \texttt{survival} & \cite{prentice1973}  & \\
         \texttt{nwtco} & \texttt{survival} & \cite{breslow1999}  & \\
         \hline
         \texttt{GSE4335} & \texttt{survJamda.data} & \cite{GSE4335,yasrebi2009}  & \cite{survJamdadata} \\
         \texttt{GSE3143} & \texttt{survJamda.data} & \cite{GSE3143}  & \\
         \texttt{GSE1992} & \texttt{survJamda.data} & \cite{GSE1992}  & \\
         \hline
         \texttt{wpbc} & \texttt{TH.data} & \cite{street1995}  & \cite{THdata} \\ 
         \hline
         \texttt{TRACE} & \texttt{timereg} & \cite{jensen1997}  & \cite{timereg} \\ 
         \texttt{csl} & \texttt{timereg} & \cite{schlichting1983}  & \\ 
         \texttt{diabetes} & \texttt{timereg} & \cite{huster1989}  & \\
         \hline
         \texttt{support2} & \texttt{vanderbilt} & \cite{knaus1995}  & \cite{vanderbilt} \\ 
         \texttt{prostate} & \texttt{vanderbilt} & \cite{andrews1985}  & \\ 
         \texttt{Framingham} & \texttt{vanderbilt} & \cite{mahmood2014}  & \\
         \texttt{rhc} & \texttt{vanderbilt} & \cite{connors1996}  & \\
         \texttt{acath} & \texttt{vanderbilt} & \cite{acath}  & \\
         \texttt{wlbw} & \texttt{vanderbilt} & \cite{oshea1992}  & \\
         \hline\hline
    \end{longtable}
\end{center}

\newpage
\medskip
\printbibliography

\end{document}